\documentclass[final]{cvpr}

\usepackage{times}
\usepackage{epsfig}
\usepackage{graphicx}
\usepackage{amsmath}
\usepackage{amssymb}

\usepackage{booktabs}
\usepackage{tabularx}
\usepackage{enumerate}
\usepackage{float}
\usepackage[symbol]{footmisc}

\usepackage[font=small,labelfont=bf,tableposition=top]{caption}
\usepackage{cuted}
\usepackage{capt-of}

\usepackage[pagebackref=true,breaklinks=true,colorlinks,bookmarks=false]{hyperref}

\def\cvprPaperID{****} 
\def\confYear{CVPR 2021}

\begin{document}

\title{Automatic Image Colorization with Convolutional Neural Networks and Generative Adversarial Networks}



 \author{
 Changyuan Qiu\thanks{All authors have equal authorship and equal contribution, ranked in alphabetic order. First version of this paper was completed in 2021.} \kern15pt Hangrui Cao\footnotemark[1] \kern15pt Qihan Ren\footnotemark[1] \kern15pt Ruiyu Li\footnotemark[1] \kern15pt Yuqing Qiu\footnotemark[1]  \\
   University of Michigan\\
 	{\tt\small \{peterqiu,hangrui,qihanren,ruiyuli,qyuqing\}@umich.edu}\vspace{-0.5in}
 }



\maketitle

\begin{strip}\centering
\includegraphics[width=\textwidth]{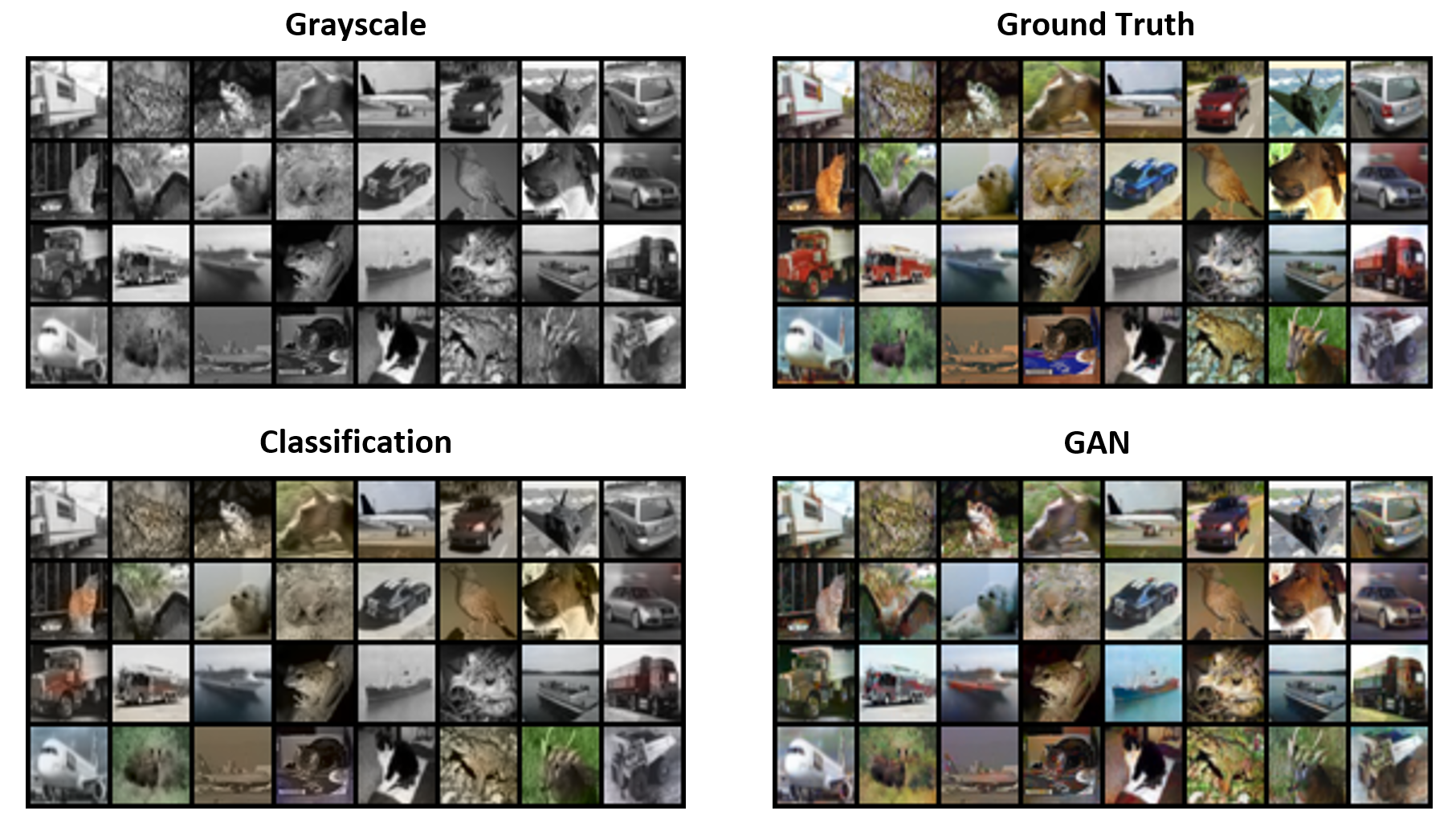}
\captionof{figure}{Colorization results on CIFAR10
\label{fig:front}}
\end{strip}




\section{Introduction}
\label{sec:introduction}


Image colorization, the task of adding colors to grayscale images, has been the focus of significant research efforts in computer vision in recent years for its various application areas such as color restoration and automatic animation colorization \cite{Nazeri_2018, anwar2020image}. The colorization problem is challenging as it is highly ill-posed with two out of three image dimensions lost, resulting in large degrees of freedom. However, semantics of the scene as well as the surface texture could provide important cues for colors: the sky is typically blue, the clouds are typically white and the grass is typically green, and there are huge amounts of training data available for learning such priors since any colored image could serve as a training data point \cite{10.1007/978-3-319-46487-9_40}. 

Colorization is initially formulated as a regression task \cite{cheng2015deep}, which ignores the multi-modal nature of color prediction. 
In this project, we explore automatic image colorization via classification and adversarial learning. We will build our models on prior works, apply modifications for our specific scenario and make comparisons.

\section{Related Works}
\label{sec:related}

Recently, deep learning techniques progressed notably for image colorization, and the fully-automatic colorization task (which does not take interactive input compared with the Scribble-based user-guided one) is commonly tackled with 3 approaches: regression, classification and adversarial learning \cite{anwar2020image, cheng2015deep,  10.1007/978-3-319-46487-9_40,  isola2017image, Nazeri_2018}.

\cite{cheng2015deep} takes the lead in investigating fully-automatic colorization with deep neural networks (DNNs) and formulates image colorization as a regression problem: the DNN takes the extracted feature descriptors at each pixel as input and outputs the continuous values of the colored channels at the corresponding pixel, with the loss function to be the Mean Square Error (MSE) between the predicted colored channel values and the ground truth. \cite{10.1007/978-3-319-46487-9_40} first formulates the task as a classification problem by quantizing the $ab$ channels of the CIE  $Lab$ color space into 313 discrete $ab$ pairs and using the multinomial cross entropy loss to train the model. \cite{isola2017image, Nazeri_2018} uses conditional Generative Adversarial Networks (cGANs) for automatic colorization of grayscale images.
\section{Approach}
\label{sec:approach}
We summarize the overall objective of colorization as follows: given a single lightness channel of an image $X\in \mathbb{R}^{H \times W \times 1}$, predict the two corresponding color channels $\hat Y=\mathcal{F}(X)\in \mathbb{R}^{H \times W \times 2}$, where $\mathcal{F}$ is the mapping function to be learned.

\begin{figure*}
\begin{center}
\includegraphics[width=\linewidth]{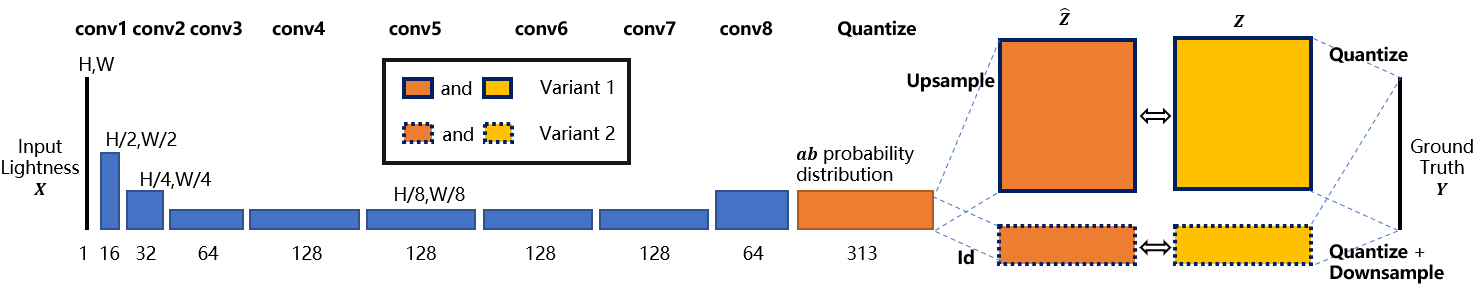}
\end{center}
\vspace{-0.2in}
\caption{Network architecture for classification-based colorization. Each Conv block is composed of 2 or 3 repeated Conv and ReLU layers. All spatial downsamplings in the network are achieved by Conv layer with stride greater than 1. We have two variants for this architecture w.r.t the output layer. \textbf{Variant 1:} we upsample the ab probability distribution to the original size, and calculate loss w.r.t the quantized ground truth. The upsampling methods include bilinear interpolation and ConvTranspose. \textbf{Variant 2:} we keep the \textit{ab} probability distribution unchanged (\textbf{Id} in the figure) , but downsample the quantized ground truth to make the size match.}
\label{fig:short1}
\end{figure*}

\begin{figure*}
\begin{center}
\includegraphics[width=\linewidth]{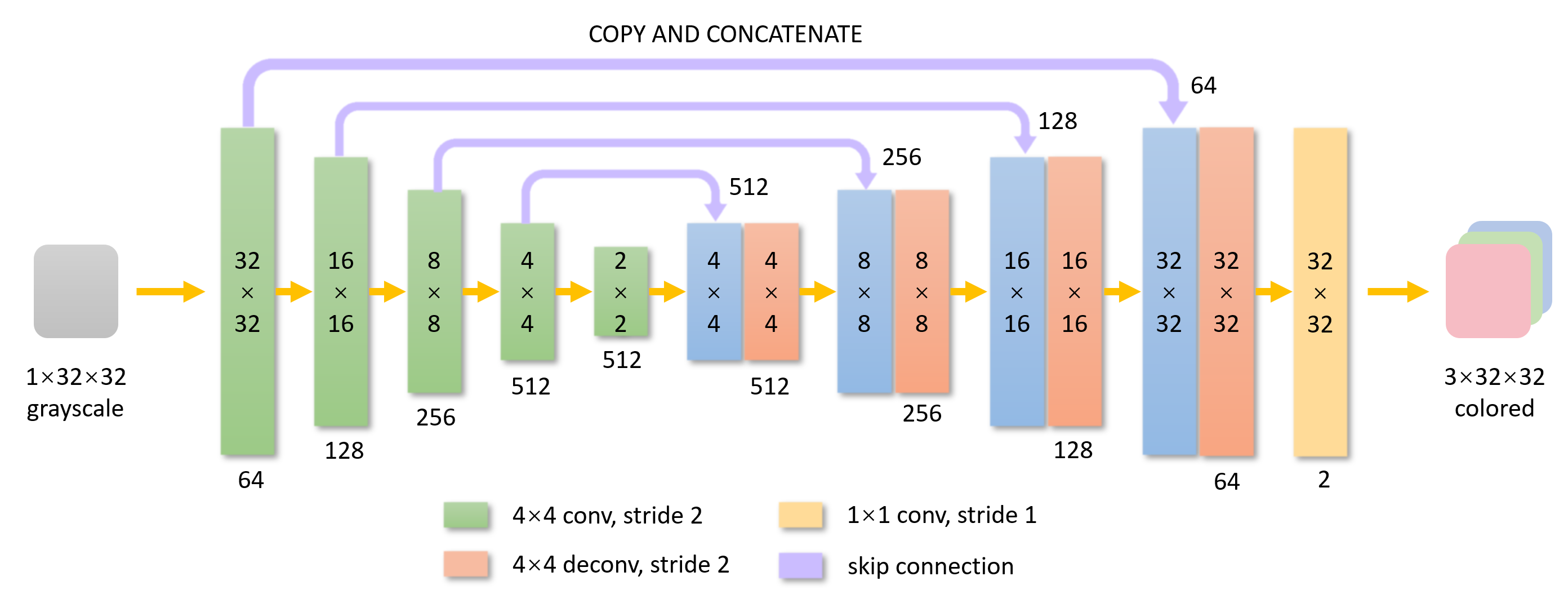}
\end{center}
\vspace{-0.2in}
\caption{Network architecture for generator (U-Net)  of GAN. The symmetric architecture consists of left contracting path and right expansive path. For the green contracting path (encoder), each block is a $4\times4$ convolutional layer with stride 2 for down-sampling, followed by batch normalization and Leaky-ReLU activation function with slope 0.2. For the right expansive path (decoder), each block consists of a $4\times4$ transposed convolutional layer with stride 2 for up-sampling and concatenation with the mirroring layer from the contracting path. Then, the concatenated block goes through a $3\times 3$ convolution layer with stride 1 for halving channels, followed by batch normalization and ReLU activation function. The output layer is a $1\times1$ convolution layer followed by a Tanh activation function. }
\vspace{-0.1in}
\label{fig:short2}
\end{figure*}




\subsection{Classification-based Colorization}
In this approach, we treat the colorization task as a classification problem instead of a regression problem due to the ambiguity of colorization.
  We employ an architecture similar to U-Net~\cite{ronneberger2015unet} without skip-connection and with  dilated convolutions following ~\cite{10.1007/978-3-319-46487-9_40, chen2017deeplab, YuKoltun2016}. Details of our architecture are shown in Figure \ref{fig:short1} with two variants.
  
  \par
  To formalize, the \textit{ab} color space is quantized to bins of size 10, and yields a total of 313 possible \textit{ab} pairs. Our network maps the input $X$ to a probability distribution $\hat Z\in [0,1]^{H \times W \times Q}$, where $Q=313$ is the number of quantized \textit{ab} pairs. Our loss function is formulated as follows.
{\small 
\setlength{\belowdisplayskip}{0em}
\[
    L_{cl}(\hat{Z}, Z)=-\sum_{h,w}\sum_q Z_{h,w,q}\log(\hat{Z}_{h,w,q})\]
    
    }
 where $\hat{Z}$ is the predicted probability distribution and $Z$ is the quantized ground truth, using the soft-encoding scheme in ~\cite{10.1007/978-3-319-46487-9_40}. Therefore, the network is not strictly end-to-end learned. To obtain the final colorized image, we first map the predicted probability distribution $\hat Z$ to color channels $\hat Y$ by function $\hat Y = \mathcal{H} (\hat Z)$, and then concatenate $\hat Y$ to the input lightness channel $X$. For function $\mathcal{H}$, we use the \textit{annealed-mean} operation in ~\cite{10.1007/978-3-319-46487-9_40} with temperature $T=0.38$.
  
  \par
 Unlike~\cite{10.1007/978-3-319-46487-9_40}, we did not apply class rebalancing. The technique is originally used to correct the bias towards lower \textit{ab} values, but we empirically observed that it disrupted the training process in our case.

\subsection{GAN-based Colorization}
Generative Adversarial Networks (GAN) \cite{goodfellow2014generative} are composed of two competing neural network models. For this colorization problem setting, the generator takes grayscale images and generates colorized versions; the discriminator gets colored images either from the generator or the labels, concatenated with the grayscale images, and tries to identify which pair contains the real colored image~\cite{radford2016unsupervised}. 
Similar to the approach in \cite{radford2016unsupervised}, we utilize deep convolutional neural networks as generative models for our adversarial framework. Since for the colorization problem setting, the input is grayscale images instead of random noises, we employ a conditional GAN instead of the traditional one. The cost functions are formulated as follows.
\begin{equation*}
\footnotesize
\begin{aligned}
&\min _{\theta_{G}} J^{(G)}\left(\theta_{D}, \theta_{G}\right)=
\min _{\theta_{G}}-\mathbb{E}_{z}\left[\log \left(D\left(G\left(\mathbf{0}_{z}| x\right)\right)\right)\right]+\lambda\left\|G\left(\mathbf{0}_{z} | x\right)-y\right\|_{1} 
\end{aligned}
\end{equation*}
\begin{equation*}
\footnotesize
\begin{aligned}
&\max _{\theta_{D}} J^{(D)}\left(\theta_{D}, \theta_{G}\right)=
\max _{\theta_{D}}\left(\mathbb{E}_{y}[\log (D(y | x))]+\mathbb{E}_{z}[\log \left(1-D\left(G\left(\mathbf{0}_{z}| x\right) | x\right)\right)]\right)
\end{aligned}
\end{equation*}
where $G\left(\mathbf{0}_{z}| x\right)$ is the colorized image produced by the generator, with input as zero noise $\mathbf{0}_{z}$ with the grayscale image $x$ as a prior and $y$ is the ground truth.

We build up and train a Conditional Deep Convolutional Generative Adversarial Network (C-DCGAN) following \cite{Nazeri_2018}. We employ a modified U-Net~\cite{ronneberger2015unet} for our basic architecture. For the generator $G$, it is constructed as the modified U-Net model as shown in Figure \ref{fig:short2}. For the discriminator $D$, it only utilizes the contracting part (encoder) in the U-Net model, with the number of channels being doubled after each down-sampling. The output layer is a $4\times 4$ convolutional layer with stride 1, which generates a 1 dimensional output. Finally, the sigmoid activation function is used to map the output to a probability of the input image being real.

To better control the color space, we separate the brightness channel and color channels using the CIE $Lab$ color space, where we only need to predict two color channels $ab$ in the generator.

\section{Experiment}
\label{sec:experiment}

\paragraph*{Dataset}
\label{sec:dataset}
\par We evaluate our models on the canonical dataset CIFAR-10
\cite{Krizhevsky09learningmultiple}. CIFAR-10 consists of 60000 images of resolution $32 \times 32$ uniformly partitioned in 10 classes, with each class having 6000 images. 
Specifically, for each class, 5000 images are randomly selected for training and the remaining 1000 images are left for testing. 


\paragraph*{Training}
\label{sec:training}

We train both models using Adam \cite{Adam}, with learning rate of 1e-3 for the classfication-based model, and learning rate of 1e-4 for both generator and discriminator of the GAN-based model. The regularization term $\lambda$ is set as 100 for the generator of the GAN-based model. Our classfication-based model is trained for 100 epochs, and takes 4.5 hours to train on one Tesla V100 hosted on Google Colab Pro; our GAN-based model is trained for 200 epochs, and takes 4 hours on one Nvidia GTX 2070 hosted on a remote server.

\subsection{Evaluation Metrics}
\label{sec:evaluation}
\paragraph{Pixel-wise Accuracy}
To measure the difference between the learned images from our models and the true images, we first measure the mean absolute error between pixels. For one test image, the accuracy is measured by the ratio of the number of pixels whose errors are smaller than the error threshold $\epsilon$. Formally, denote $pred(i,j)$ as one image pixel in image generated by the model and $real(i,j)$ as the original image pixel values. Both pixel values are normalized to $[0,1]$. If the absolute difference is smaller than $\epsilon$, namely
$$
    |pred(i, j) - real(i,j)| < \epsilon,
$$

then $pred(i,j)$ will be classified as true in terms of color information compared with the pixel in original ones. As each pixel has R, G, B values, we mark one pixel as true when all of its three channel values satisfy the above equation.

        

        
\paragraph*{PSNR and SSIM}
Besides, we use peak-signal-to-noise ratio (PSNR) and structrural similarity index (SSIM index) to measure image qualities \cite{Metricss}. PSNR is defined in log form of the mean sqaure error and we calculate PSNR for each generated images( after transformation from $Lab$ to $rgb$). SSIM is a method to measure the similarity between images. We calculate average PSNR and SSIM values on the test dataset and the results are shown below.

         


\begin{table*}[!htbp]
    \centering
    \begin{tabularx}{1\textwidth}{lcccc}
    \toprule
      \textbf{Model}                                    & Pixel-Acc\ \textbf{$\epsilon = 2\%$} & Pixel-Acc\ \textbf{ $\epsilon = 5\%$}&PSNR (dB) & SSIM \\ \hline\hline
    Classification (Upsample - Bilinear Interpolation ) & 0.888\%  &  5.272\% &   21.491  & 0.913  \\ 
    Classification (Upsample - Deconv) & 0.919\%  & 5.189\%  & 21.220  & 0.908  \\
     Classification (Downsample) & 0.923\% & 5.828\% & 21.848 & 0.913   \\ 
      GAN (ours)  & 33.255\% & 57.510\% & 24.608 & 0.910\\  \hline
     GAN \cite{Nazeri_2018}  & 24.100\%& 65.500\% & — &— \\    
    \bottomrule
    \end{tabularx}

        \caption{Model Performance Comparison on CIFAR-10\label{quan1}}
            \vspace{-0.3in}
    \end{table*}

\subsection{CIFAR-10 Results}
\label{sec:result}

\begin{figure}
    \centering
    \includegraphics[width=0.3\textwidth]{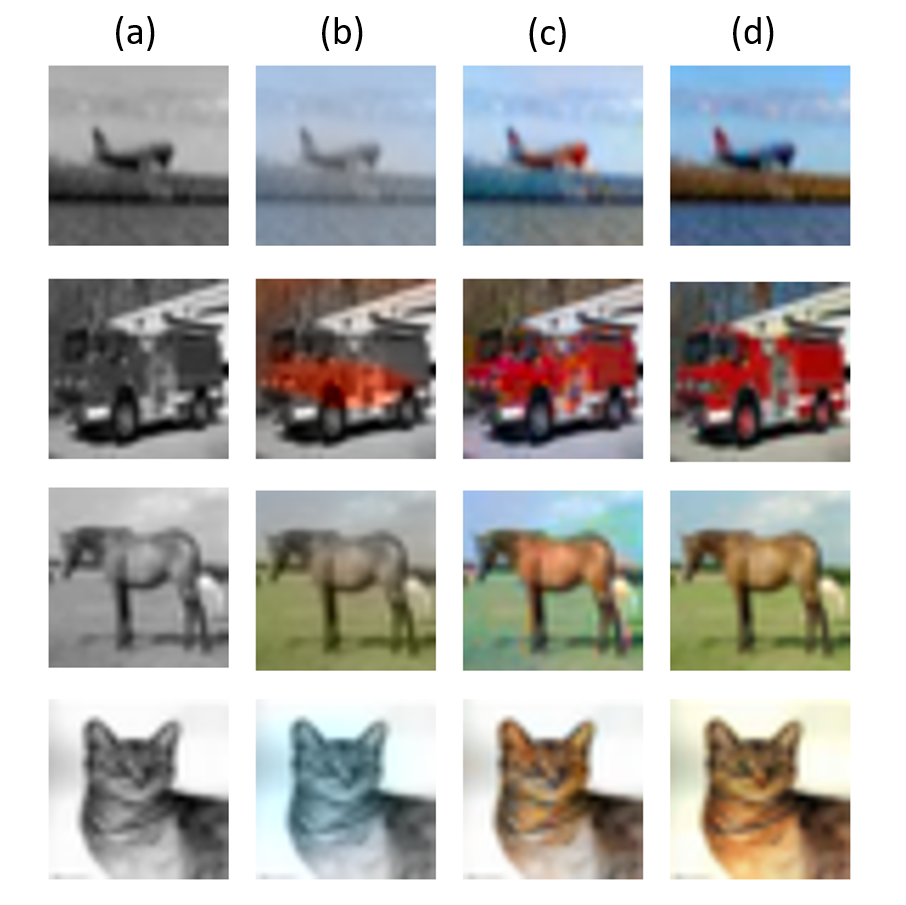}
    \caption{Comparison of Colarization Results on CIFAR-10. \\ (a) Grayscale. (b) Classification. (c) GAN. (d) Ground Truth.\label{fig:fig_comp}}
    \vspace{-0.15in}
\end{figure}

We carried out experiments on the CIFAR-10 dataset. Both the classification method and the GAN method are able to automatically colorize grayscale images to an acceptable visual degree. Qualititative results are shown in Figure \ref{fig:front}, \ref{fig:fig_comp}. Quantatitive results are shown in Table \ref{quan1}. Compared with classficiation methods, images generated by GAN has much higher pixel-wise accuracy and higher PSNR(dB) values, indicating that in general the GAN method performs better than classification. The SSIM values observed by two methods are approximately the same. As we calculate the metrics based on the three channels (R, G, B), we compare the pixel-wise accuracy of three channels and find that for both two methods, the accuracy in R channel is lower than the other two channels, which implies our models are weak to generate colors in R channels.

\subsection{User Study}
\label{sec:user}
We carried out a fool study process by randomly picking 200 generated sets, where each set has three images, the ground truth and colorized images learned by classification and by GAN respectively. We asked 16 students who did not know the image labels in advance to identify which image is the ground truth in each set. The identified class ratios are listed in table \ref{userstud}. We also asked users to rate images ranging from 1 - 5, with higher score indicating better image reality and quality. 
 \vspace{-0.1in}
\begin{table}[H]
    \centering
    \small
    \begin{tabular}{cccc}
    \hline
      Identified class ratio &  Ground Truth & Classification  & GAN \\
      \hline
      \#    identified / \#total   &  54.91\% & 4.80\% & 40.69 \%  \\
      Avg Score & 4.0 & 2.3 & 3.7 \\
      \hline
    \end{tabular}
    \caption{User Study Table}
    \label{userstud}
    \vspace{-0.2in}
\end{table}
    \vspace{-0.1in}
The above results show that images generated by GAN and CNN can fool users to some extent.

\section{Implementation}
\label{sec:implementation}
For classification-based approach, we adopt the architecture in ~\cite{10.1007/978-3-319-46487-9_40} but reduce the number of channels by a factor of $\frac{1}{4}$ to align with the smaller image size $(32\times 32)$ in CIFAR-10. In addition, we implement two variants of the architecture, with different downsampling or upsampling methods for the output layer (see Figure \ref{fig:short1}). We implement the training and evaluation code from scratch in PyTorch, and tune the hyperparameters by ourselves. We utilize the tensorboard function in PyTorch to display synchronous colorization results.  The code snippets for color conversion and quantization of \textit{ab} color space are borrowed from \cite{zhang2017real} and \cite{Bing2020} respectively.

For GAN-based approach, we adopt the architecture of C-DCGAN and hyperparameters given in \cite{Nazeri_2018} but reduce the number of layers in the U-Net basic architecture as we are dealing with smaller images (see Figure \ref{fig:short2}). We implement the models of generator and discriminator in PyTorch from scratch and simplify the code to make it more straightforward. We utilize the tensorboard function in PyTorch to display synchronous colorization results. We refer to dataloader code, training code and color conversion and quantization code from \textit{rgb} to \textit{Lab} in \cite{pytorchgan} but we implement our own evaluation code, color conversion and quantization code from \textit{Lab} to \textit{rgb}.

\section{Conclusion}
In this project, we compare and evaluate the performance of convolutional neural networks and generative adversarial networks on automatic image colorization tasks. Both of them are able to automatically colorize grayscale images to an acceptable  visual degree. Compared with the classification-based CNN method, C-DCGAN perform much better while is also more computationally expensive at the same time.
\section{Future Work}
\label{sec:future}
We plan to experiment with images of higher resolutions from dataset like ImageNet \cite{5206848} $(224\times 224)$  or MS COCO \cite{lin2014microsoft}  $(640\times 480)$. Besides, modifying the backbone of the classifier (like change to ResNet \cite{he2016deep}) can potentially improve the performance. We also plan to further explore other generative models like VAE \cite{kingma2013auto} and VQ-VAE \cite{oord2017neural}.


\newpage

{\small
\bibliographystyle{ieee_fullname}
\bibliography{egbib}

\begin{thebibliography}{10}\itemsep=-1pt

\bibitem{anwar2020image}
Saeed Anwar, Muhammad Tahir, Chongyi Li, Ajmal Mian, Fahad~Shahbaz Khan, and
  Abdul~Wahab Muzaffar.
\newblock Image colorization: A survey and dataset.
\newblock {\em arXiv preprint arXiv:2008.10774}, 2020.

\bibitem{pytorchgan}
Harshit Bansal.
\newblock Image colorization.
\newblock \url{https://github.com/harshitbansal05/Image-Colorization}, 2018.

\bibitem{Bing2020}
BingWin789.
\newblock colorization-traininglayers-tf.
\newblock \url{https://github.com/BingWin789/colorization-traininglayers-tf},
  2020.

\bibitem{chen2017deeplab}
Liang-Chieh Chen, George Papandreou, Iasonas Kokkinos, Kevin Murphy, and
  Alan~L. Yuille.
\newblock Deeplab: Semantic image segmentation with deep convolutional nets,
  atrous convolution, and fully connected crfs, 2017.

\bibitem{cheng2015deep}
Zezhou Cheng, Qingxiong Yang, and Bin Sheng.
\newblock Deep colorization.
\newblock In {\em Proceedings of the IEEE International Conference on Computer
  Vision}, pages 415--423, 2015.

\bibitem{5206848}
J. {Deng}, W. {Dong}, R. {Socher}, L. {Li}, {Kai Li}, and {Li Fei-Fei}.
\newblock Imagenet: A large-scale hierarchical image database.
\newblock In {\em 2009 IEEE Conference on Computer Vision and Pattern
  Recognition}, pages 248--255, 2009.

\bibitem{goodfellow2014generative}
Ian~J Goodfellow, Jean Pouget-Abadie, Mehdi Mirza, Bing Xu, David Warde-Farley,
  Sherjil Ozair, Aaron Courville, and Yoshua Bengio.
\newblock Generative adversarial networks.
\newblock {\em arXiv preprint arXiv:1406.2661}, 2014.

\bibitem{he2016deep}
Kaiming He, Xiangyu Zhang, Shaoqing Ren, and Jian Sun.
\newblock Deep residual learning for image recognition.
\newblock In {\em Proceedings of the IEEE conference on computer vision and
  pattern recognition}, pages 770--778, 2016.

\bibitem{Metricss}
A. {Horé} and D. {Ziou}.
\newblock Image quality metrics: Psnr vs. ssim.
\newblock In {\em 2010 20th International Conference on Pattern Recognition},
  pages 2366--2369, 2010.

\bibitem{isola2017image}
Phillip Isola, Jun-Yan Zhu, Tinghui Zhou, and Alexei~A Efros.
\newblock Image-to-image translation with conditional adversarial networks.
\newblock In {\em Proceedings of the IEEE conference on computer vision and
  pattern recognition}, pages 1125--1134, 2017.

\bibitem{Adam}
Diederik~P. Kingma and Jimmy Ba.
\newblock Adam: A method for stochastic optimization, 2017.

\bibitem{kingma2013auto}
Diederik~P Kingma and Max Welling.
\newblock Auto-encoding variational bayes.
\newblock {\em arXiv preprint arXiv:1312.6114}, 2013.

\bibitem{Krizhevsky09learningmultiple}
Alex Krizhevsky.
\newblock Learning multiple layers of features from tiny images.
\newblock 2009.

\bibitem{lin2014microsoft}
Tsung-Yi Lin, Michael Maire, Serge Belongie, James Hays, Pietro Perona, Deva
  Ramanan, Piotr Doll{\'a}r, and C~Lawrence Zitnick.
\newblock Microsoft coco: Common objects in context.
\newblock In {\em European conference on computer vision}, pages 740--755.
  Springer, 2014.

\bibitem{Nazeri_2018}
Kamyar Nazeri, Eric Ng, and Mehran Ebrahimi.
\newblock Image colorization using generative adversarial networks.
\newblock In {\em International conference on articulated motion and deformable
  objects}, pages 85--94. Springer, 2018.

\bibitem{oord2017neural}
Aaron van~den Oord, Oriol Vinyals, and Koray Kavukcuoglu.
\newblock Neural discrete representation learning.
\newblock {\em arXiv preprint arXiv:1711.00937}, 2017.

\bibitem{radford2016unsupervised}
Alec Radford, Luke Metz, and Soumith Chintala.
\newblock Unsupervised representation learning with deep convolutional
  generative adversarial networks, 2016.

\bibitem{ronneberger2015unet}
Olaf Ronneberger, Philipp Fischer, and Thomas Brox.
\newblock U-net: Convolutional networks for biomedical image segmentation,
  2015.

\bibitem{YuKoltun2016}
Fisher Yu and Vladlen Koltun.
\newblock Multi-scale context aggregation by dilated convolutions.
\newblock In {\em ICLR}, 2016.

\bibitem{10.1007/978-3-319-46487-9_40}
Richard Zhang, Phillip Isola, and Alexei~A. Efros.
\newblock Colorful image colorization.
\newblock In Bastian Leibe, Jiri Matas, Nicu Sebe, and Max Welling, editors,
  {\em Computer Vision -- ECCV 2016}, pages 649--666, Cham, 2016. Springer
  International Publishing.

\bibitem{zhang2017real}
Richard Zhang, Jun-Yan Zhu, Phillip Isola, Xinyang Geng, Angela~S Lin, Tianhe
  Yu, and Alexei~A Efros.
\newblock Real-time user-guided image colorization with learned deep priors.
\newblock {\em ACM Transactions on Graphics (TOG)}, 9(4), 2017.

\end{thebibliography}
}

\end{document}